\documentclass[12pt,twoside]{article}
\usepackage{latexsym}
\def\,{\mskip 3mu} \def\>{\mskip 4mu plus 2mu minus 4mu} \def\;{\mskip 5mu plus 5mu} \def\!{\mskip-3mu}
\def\dispmuskip{\thinmuskip= 3mu plus 0mu minus 2mu \medmuskip=  4mu plus 2mu minus 2mu \thickmuskip=5mu plus 5mu minus 2mu}
\def\textmuskip{\thinmuskip= 0mu                    \medmuskip=  1mu plus 1mu minus 1mu \thickmuskip=2mu plus 3mu minus 1mu}
\textmuskip
\def\beq{\dispmuskip\begin{equation}}    \def\eeq{\end{equation}\textmuskip}
\def\beqn{\dispmuskip\begin{displaymath}}\def\eeqn{\end{displaymath}\textmuskip}
\def\bqa{\dispmuskip\begin{eqnarray}}    \def\eqa{\end{eqnarray}\textmuskip}
\def\bqan{\dispmuskip\begin{eqnarray*}}  \def\eqan{\end{eqnarray*}\textmuskip}
\newtheorem{theorem}{Theorem}
\newtheorem{corollary}[theorem]{Corollary}
\newtheorem{lemma}[theorem]{Lemma}

\newenvironment{keywords}{\centerline{\bf\small
Keywords}\begin{quote}\small}{\par\end{quote}\vskip 1ex}
\def\citet{\cite}\def\citep{\cite}\def\citealt{\cite}\def\citeauthor{\cite}
\def\myparskip{\vspace{1.5ex plus 0.5ex minus 0.5ex}\noindent}
\def\paragraph#1{\myparskip{\bfseries\boldmath{#1.}}}
\def\paradot#1{\myparskip{\bfseries\boldmath{#1.}}}
\def\paranodot#1{\myparskip{\bfseries\boldmath{#1}}}
\def\eps{\varepsilon}
\def\nq{\hspace{-1em}}
\def\qed{\hspace*{\fill}$\Box\quad$\\}
\def\odt{{\textstyle{1\over 2}}}

\def\SetR{I\!\!R}
\def\SetN{I\!\!N}

\def\D{{\cal D}}
\def\S{{\cal S}}
\def\E{{\cal E}}
\def\X{{\cal X}}                        % input/perception set/alphabet
\def\Y{{\cal Y}}                        % output/action set/alphabet
\def\qmbox#1{{\quad\mbox{#1}\quad}}
\def\scp{{\scriptscriptstyle^{\,\circ}}}
\def\sooe{{\textstyle{1\over\eta}}}
\def\FPL{\text{FPL} }
\def\IFPL{\text{IFPL} }

\def\leqt{_{1:t}}

\def\leqT{_{1:T}}
\def\ltT{_{<T}}
\def\ltt{_{<t}}

\def\smin{^{min}}

\def\leqs#1{\stackrel {#1} \leq}
\def\text#1{\mbox{\scriptsize{#1}}}
\def\e{{\rm e}}                        % natural e

\begin{document}
\title{\vskip -10mm\normalsize\sc Technical Report \hfill IDSIA-10-05
\vskip 2mm\bf\Large\hrule height5pt \vskip 6mm
Adaptive Online Prediction by \\ Following the Perturbed Leader
\vskip 6mm \hrule height2pt \vskip 5mm}
\author{{\bf Marcus Hutter} and {\bf Jan Poland}\\[3mm]
\normalsize IDSIA, Galleria 2, CH-6928\ Manno-Lugano, Switzerland%
\thanks{This work was supported by SNF grant 2100-67712.02.\newline\hspace*{3.6ex}
A shorter version appeared in the proceedings of the ALT 2004 conference \citep{Hutter:04expert}.}\\
\normalsize \{marcus,jan\}@idsia.ch, \ http://www.idsia.ch/$^{_{_\sim}}\!$\{marcus,jan\} }
\date{14 April 2005}
\maketitle

\begin{abstract}%
When applying aggregating strategies to Prediction with Expert
Advice, the learning rate must be adaptively tuned. The
natural choice of $\sqrt{\mbox{complexity/current loss}}$
renders the analysis of Weighted Majority derivatives quite
complicated. In particular, for arbitrary weights there have
been no results proven so far. The analysis of the alternative
``Follow the Perturbed Leader'' (FPL) algorithm from Kalai and
Vempala (2003) based on Hannan's algorithm is easier. We
derive loss bounds for adaptive learning rate and both finite
expert classes with uniform weights and countable expert
classes with arbitrary weights. For the former setup, our loss
bounds match the best known results so far, while for the
latter our results are new.
\end{abstract}


\newpage
%------------------------------%
%      Table of Contents       %
%------------------------------%
\begin{quote}\begin{quote}
\def\contentsname{\normalsize \hfil Contents \hfil}
{\parskip=-2.5ex\tableofcontents}
\end{quote}\end{quote}

%%%%%%%%%%%%%%%%%%%%%%%%%%%%%%%%%%%%%%%%%%%%%%%%%%%%%%%%%%%%%%%
\section{Introduction}\label{secInt}
%%%%%%%%%%%%%%%%%%%%%%%%%%%%%%%%%%%%%%%%%%%%%%%%%%%%%%%%%%%%%%%

%-------------------------------%
%\paradot{Prediction with Expert Advice}
%-------------------------------%
In Prediction with Expert Advice (PEA) one considers an
ensemble of sequential predictors (experts). A master
algorithm is constructed based on the historical performance
of the predictors. The goal of the master algorithm is to
perform nearly as well as the best expert in the class, on any
sequence of outcomes. This is achieved by making (randomized)
predictions close to the better experts.

%-------------------------------%
%\paradot{Historical Survey}
%-------------------------------%
PEA theory has rapidly developed in the recent past.
Starting with the Weighted Majority (WM) algorithm of
\citet{Littlestone:89,Littlestone:94} and the aggregating
strategy of \citet{Vovk:90}, a vast variety of different
algorithms and variants have been published. A key parameter
in all these algorithms is the \emph{learning rate}. While
this parameter had to be fixed in the early algorithms such as
WM, \citet{Cesa:97} established the so-called doubling trick
to make the learning rate coarsely adaptive. A little later,
incrementally adaptive algorithms were developed by
\citet{Auer:00,Auer:02pea,Yaroshinsky:04,Gentile:03}, and
others. In Section \ref{secConc}, we will compare our results
with these works more in detail. Unfortunately, the loss bound
proofs for the incrementally adaptive WM variants are quite
complex and technical, despite the typically simple and
elegant proofs for a static learning rate.

%-------------------------------%
%\paradot{Adaptive Learning Rate}
%-------------------------------%
The complex growing proof techniques also had another consequence:
While for the original WM algorithm, assertions are proven for
countable classes of experts with arbitrary weights, the modern
variants usually restrict to finite classes with uniform weights
(an exception being \citet{Gentile:03}, see the discussion
section). This might be sufficient for many practical purposes but
it prevents the application to more general classes of predictors.
Examples are extrapolating (=predicting) data points with the help
of a polynomial (=expert) of degree $d=1,2,3,...$ --or-- the (from
a computational point of view largest) class of all computable
predictors. Furthermore, most authors have concentrated on
predicting \emph{binary} sequences, often with the 0/1 loss for
$\{0,1\}$-valued and the absolute loss for $[0,1]$-valued
predictions. Arbitrary losses are less common. Nevertheless, it is
easy to abstract completely from the predictions and consider the
resulting losses only. Instead of predicting according to a
``weighted majority'' in each time step, one chooses one
\emph{single} expert with a probability depending on his past
cumulated loss. This is done e.g.\ by \citet{Freund:97}, where an
elegant WM variant, the Hedge algorithm, is analyzed.

%-------------------------------%
%\paradot{Follow the Perturbed Leader}
%-------------------------------%
A different, general approach to achieve similar results is
``Follow the Perturbed Leader'' (FPL). The principle dates
back to as early as 1957, now called Hannan's algorithm
\citep{Hannan:57}. In 2003, Kalai and Vempala published a
simpler proof of the main result of Hannan and also succeeded
to improve the bound by modifying the distribution of the
perturbation\nocite{Kalai:03}. The resulting algorithm (which
they call FPL*) has the same performance guarantees as the
WM-type algorithms for fixed learning rate, save for a factor
of $\sqrt 2$. A major advantage we will discover in this work
is that its analysis remains easy for an adaptive learning
rate, in contrast to the WM derivatives. Moreover, it
generalizes to online decision problems other than PEA.

%-------------------------------%
%\paradot{What' new}
%-------------------------------%
In this work, we study the FPL algorithm for PEA. The problems of
WM algorithms mentioned above are addressed: Bounds on the
cumulative regret of the standard form $\sqrt{kL}$ (where $k$ is
the complexity and $L$ is the cumulative loss of the best expert
in hindsight) are shown for countable expert classes with
arbitrary weights, adaptive learning rate, and arbitrary losses.
Regarding the adaptive learning rate, we obtain proofs that are
simpler and more elegant than for the corresponding WM algorithms.
(In particular, the proof for a self-confident choice of the
learning rate, Theorem~\ref{thFPLLDynamic}, is less than half a
page.) Further, we prove the first loss bounds for \emph{arbitrary
weights} and adaptive learning rate. In order to obtain the
optimal $\sqrt{kL}$ bound in this case, we will need to introduce
a hierarchical version of FPL, while without hierarchy we show a
worse bound $k\sqrt{L}$. (For self-confident learning rate
together with uniform weights and arbitrary losses, one can prove
corresponding results for a variant of WM by adapting an argument
by \citealt{Auer:02pea}.)

%-------------------------------%
%\paradot{Online, worst case and probabilities}
%-------------------------------%
PEA usually refers to an \emph{online worst case} setting: $n$
experts that deliver sequential predictions over a time range
$t=1,\ldots,T$ are given. At each time $t$, we know the actual
predictions and the \emph{past} losses. The goal is to give a
prediction such that the overall loss after $T$ steps is ``not
much worse'' than the best expert's loss \emph{on any sequence of
outcomes}. If the prediction is deterministic, then an adversary
could choose a sequence which provokes maximal loss. So we have to
\emph{randomize} our predictions. Consequently, we ask for a
prediction strategy such that the \emph{expected} loss on any
sequence is small.

%-------------------------------%
%\paradot{Contents}
%-------------------------------%
This paper is structured as follows. In Section~\ref{secSetup}
we give the basic definitions. While \citeauthor{Kalai:03}
consider general online decision problems in
finite-dimensional spaces, we focus on online prediction tasks
based on a countable number of experts. Like \citet{Kalai:03}
we exploit the infeasible FPL predictor (IFPL) in our
analysis.
Sections~\ref{secIFPL} and \ref{secFFPL} derive the main
analysis tools. In Section~\ref{secIFPL} we generalize (and
marginally improve) the upper bound \citep[Lem.3]{Kalai:03} on
IFPL to arbitrary weights. The main difficulty we faced was to
appropriately distribute the weights to the various terms. For
the corresponding lower bound (Section~\ref{secLowFPL}) this
is an open problem.
In Section~\ref{secFFPL} we exploit our restricted setup to
significantly improve \citep[Eq.(3)]{Kalai:03} allowing for
bounds logarithmic rather than linear in the number of
experts.
The upper and lower bounds on IFPL are combined to derive
various regret bounds on FPL in Section~\ref{secBounds}.
Bounds for static and dynamic learning rate in terms of the
sequence length follow straight-forwardly. The proof of our
main bound in terms of the loss is much more elegant than the
analysis of previous comparable results.
Section~\ref{secHierarchy} proposes a novel hierarchical procedure
to improve the bounds for non-uniform weights.
In Section~\ref{secLowFPL}, a lower bound is established.
In Section~\ref{secAdap}, we consider the case of independent
randomization more seriously. In particular, we show that the
derived bounds also hold for an adaptive adversary.
Section~\ref{secMisc} treats some additional issues, including
bounds with high probability, computational aspects, deterministic
predictors, and the absolute loss.
Finally, in Section~\ref{secConc} we discuss our results, compare
them to references, and state some open problems.

%%%%%%%%%%%%%%%%%%%%%%%%%%%%%%%%%%%%%%%%%%%%%%%%%%%%%%%%%%%%%%%
\section{Setup and Notation}\label{secSetup}
%%%%%%%%%%%%%%%%%%%%%%%%%%%%%%%%%%%%%%%%%%%%%%%%%%%%%%%%%%%%%%%

%-------------------------------%
\paradot{Setup}
%-------------------------------%
Prediction with Expert Advice proceeds as follows. We are asked to
perform sequential predictions $y_t\in\Y$ at times $t=1,2,\ldots$.
At each time step $t$, we have access to the predictions
$(y_t^i)_{1\leq i\leq n}$ of $n$ experts $\{e_1,...,e_n\}$, where
the size of the expert pool is $n\in\SetN\cup\{\infty\}$. It is
convenient to use the same notation for finite ($n\in\SetN$) and
countably infinite ($n=\infty$) expert pool. After having made a
prediction, we make some observation $x_t\in\X$, and a Loss is
revealed for our and each expert's prediction. (E.g.\ the loss
might be 1 if the expert made an erroneous prediction and 0
otherwise. This is the 0/1 loss.) Our goal is to achieve a total
loss ``not much worse" than the best expert, after $t$ time steps.

We admit $n\in\SetN\cup\{\infty\}$ experts, each of which is
assigned a known complexity $k^i\geq 0$. Usually we require
$\sum_i\e^{-k^i}\leq 1$, which implies that the $k^i$ are valid
lengths of prefix code words, for instance $k^i=\ln n$ if
$n<\infty$ or $k^i=\odt+2\ln i$ if $n=\infty$. Each complexity
defines a weight by means of $\smash{\e^{-k^i}}$ and vice versa.
In the following we will talk of complexities rather than of
weights. If $n$ is finite, then usually one sets $k^i= \ln n$ for
all $i$; this is the case of \emph{uniform complexities/weights}.
If the set of experts is countably infinite ($n=\infty$), uniform
complexities are not possible. The vector of all complexities is
denoted by $k=(k^i)_{1\leq i\leq n}$. At each time $t$, each
expert $i$ suffers a loss\footnote{The setup, analysis and results
easily scale to $s_t^i\in[0,S]$ for $S>0$ other than 1.}
$s_t^i=$Loss$(x_t,y_t^i)\in[0,1]$, and $s_t=(s_t^i)_{1\leq i\leq
n}$ is the vector of all losses at time $t$. Let
$s\ltt=s_1+\ldots+s_{t-1}$ (respectively $s\leqt=s_1+\ldots+s_t$)
be the total past loss vector (including current loss $s_t$) and
$s\leqt\smin=\min_i\{s\leqt^i\}$ be the loss of the \emph{best
expert in hindsight (BEH)}. Usually we do not know in advance the
time $t\geq 0$ at which the performance of our predictions are
evaluated.

%-------------------------------%
\paradot{General decision spaces}
%-------------------------------%
The setup can be generalized as follows. Let $\S\subset\SetR^n$ be the
\emph{state space} and $\D\subset\SetR^n$ the \emph{decision
space}. At time $t$ the state is $s_t\in\S$, and a decision
$d_t\in\D$ (which is made before the state is revealed) incurs a
loss $d_t\!\scp s_t$, where ``$\scp$" denotes the inner product. This
implies that the loss function is \emph{linear} in the states.
Conversely, each linear loss function can be represented in this
way. The decision which minimizes the loss in state $s\in\S$ is
\beq\label{Mdef}
  M(s):=\arg\min_{d\in\D} \{d\scp s\}
\eeq
if the minimum exists. The application of this general framework
to PEA is straightforward: $\D$ is identified with the space of
all unit vectors $\E=\{e_i:1\leq i\leq n\}$, since a decision
consists of selecting a single expert, and $s_t\in[0,1]^n$, so
states are identified with losses. Only Theorems~\ref{thIFPL} and
\ref{thLowFPL} will be stated in terms of general decision space.
Our main focus is $\D=\E$. (Even for this special case, the scalar
product notation is not too heavy, but will turn out to be
convenient.) All our results generalize to the simplex
$\D=\Delta=\{v\in[0,1]^n:\sum_i v^i=1\}$, since the minimum of a
linear function on $\Delta$ is always attained on $\E$.

%-------------------------------%
\paradot{Follow the Perturbed Leader}
%-------------------------------%
Given $s\ltt$ at time $t$, an immediate idea to solve the expert
problem is to ``Follow the Leader'' (FL), i.e.\ selecting the
expert $e_i$ which performed best in the past (minimizes
$s\ltt^i$), that is predict according to expert $M(s\ltt)$. This
approach fails for two reasons. First, for $n=\infty$ the minimum
in (\ref{Mdef}) may not exist. Second, for $n=2$ and
$s={\,0\,1\,0\,1\,0\,1 \ldots \choose \frac{1}{2}0\,1\,0\,1\,0
\ldots}$, FL always chooses the wrong prediction \citep{Kalai:03}.
We solve the first problem by penalizing each expert by its
complexity, i.e.\ predicting according to expert $M(s\ltt+k)$. The
\emph{FPL (Follow the Perturbed Leader)} approach solves the
second problem by adding to each expert's loss $s\ltt^i$ a random
perturbation.
We choose this perturbation to be negative \emph{exponentially
distributed}, either independent in each time step or once and for
all at the very beginning at time $t=0$. The former choice is
preferable in order to protect against an adaptive adversary who
generates the $s_t$, and in order to get bounds with high
probability (Section~\ref{secMisc}). For the main analysis
however, the latter choice is more convenient. Due to linearity of
expectations, these two possibilities are equivalent when dealing
with {\it expected losses} (this is straightforward for oblivious
adversary, for adaptive adversary see Section~\ref{secAdap}), so
we can henceforth assume without loss of generality one initial
perturbation $q$.

%-------------------------------%
\paranodot{The FPL algorithm} is defined as follows:\\
%-------------------------------%
%
\hspace*{1cm}Choose random vector $q\stackrel{d.}{\sim}\exp$,
             i.e.\ $P[q^1...q^n]=\e^{-q^1}\cdot...\cdot\e^{-q^n}$ for $q\geq 0$.\\
\hspace*{1cm}For $t=1,...,T$\\
\hspace*{1cm}- Choose learning rate $\eta_t$.\\
\hspace*{1cm}- Output prediction of expert $i$ which minimizes $s_{<t}^i+(k^i-q^i)/\eta_t$.\\
\hspace*{1cm}- Receive loss $s_t^i$ for all experts $i$.

\vspace{1.5ex}\noindent Other than $s\ltt$, $k$ and $q$, FPL
depends on the \emph{learning rate} $\eta_t$. We will give choices
for $\eta_t$ in Section~\ref{secBounds}, after having established
the main tools for the analysis. The expected loss at time $t$ of
FPL is $\ell_t:=E\big[M(s_{<t}+{k-q\over\eta_t})\scp s_t\big]$.
The key idea in the FPL analysis is the use of an intermediate
predictor \emph{IFPL} (for \emph{Implicit or Infeasible FPL}).
IFPL predicts according to $M(s\leqt+\smash{k-q\over\eta_t})$,
thus under the knowledge of $s_t$ (which is of course not
available in reality). By
$r_t:=E\big[M(s_{1:t}+\smash{k-q\over\eta_t})\scp s_t\big]$ we
denote the expected loss of IFPL at time $t$. The losses of IFPL
will be upper-bounded by BEH in Section~\ref{secIFPL} and
lower-bounded by FPL in Section~\ref{secFFPL}. Note that our
definition of the FPL algorithm deviates from that of
\citeauthor{Kalai:03}. It uses an exponentially distributed
perturbation similar to their FPL$^*$ but one-sided and a
non-stationary learning rate like Hannan's algorithm.

%-------------------------------%
\paradot{Notes}
%-------------------------------%
Observe that we have stated the FPL algorithm regardless of the
actual \emph{predictions} of the experts and possible
\emph{observations}, only the \emph{losses} are relevant.
Note also that an expert can implement a highly complicated strategy
depending on past outcomes, despite its trivializing
identification with a constant unit vector. The complex expert's
(and environment's) behavior is summarized and hidden in the state
vector $s_t=$Loss$(x_t,y_t^i)_{1\leq i\leq n}$.
Our results therefore apply to \emph{arbitrary prediction and
observation spaces $\Y$ and $\X$ and arbitrary bounded loss
functions}.
This is in contrast to the major part of PEA work
developed for binary alphabet and 0/1 or absolute loss only.
Finally note that the setup allows for losses generated by an
adversary who tries to maximize the regret of FPL and knows the
FPL algorithm and all experts' past predictions/losses. If the
adversary also has access to FPL's past decisions, then FPL must
use independent randomization at each time step in order to
achieve good regret bounds.

%-------------------------------%
\paradot{Motivation of FPL}
%-------------------------------%
Let $d(s_{<t})$ be any predictor with decision based on $s_{<t}$.
The following identity is easy to show:
\beq\label{eqFId}
  \underbrace{\sum_{t=1}^T d(s_{<t})\scp s_t}_{\text{``FPL''}}
  \;\equiv\;
  \underbrace{_{\rule{0ex}{3.8ex}}d(s_{1:T})\scp s_{1:T}}_{\text{``BEH''}}
  + \overbrace{\underbrace{\sum_{t=1}^T [d(s_{<t})\!-\!d(s_{1:t})]\scp s_{<t}}_{\text{``IFPL}-\text{BEH''}}}^{\text{$\leq 0$ if $d\approx M$}}
  + \overbrace{\underbrace{\sum_{t=1}^T [d(s_{<t})\!-\!d(s_{1:t})]\scp s_t}_{\text{``FPL}-\text{IFPL''}}}^{\text{small if $d(\cdot)$ is continuous}}
\eeq
For a good bound of FPL in terms of BEH we need the first term on
the r.h.s.\ to be close to BEH and the last two terms to be small.
The first term is close to BEH if $d\approx M$. The second to last
term is even negative if $d=M$, hence small if $d\approx M$. The
last term is small if $d(s_{<t})\approx d(s_{1:t})$, which is the
case if $d(\cdot)$ is a sufficiently smooth function.
Randomization smoothes the discontinuous function $M$: The
function $d(s):=E[M(s-q)]$, where $q\in\SetR^n$ is some random
perturbation, is a continuous function in $s$. If the mean and
variance of $q$ are small, then $d\approx M$, if the variance of
$q$ is large, then $d(s_{<t})\approx d(s_{1:t})$. An intermediate
variance makes the last two terms of (\ref{eqFId}) simultaneously
small enough, leading to excellent bounds for FPL.

%-------------------------------%
\paradot{List of notation}\hfill\\
%-------------------------------%
$n\in\SetN\cup\{\infty\}$ ($n=\infty$ means countably infinite $\E$).\\
$x^i$ is $i$th component of vector $x\in\SetR^n$.\\
$\E:=\{e_i:1\leq i\leq n\}=$ set of unit vectors ($e_i^j=\delta_{ij}$).\\
$\Delta:=\{v\in[0,1]^n:\sum_i v^i=1\}$= simplex.\\
$s_t\in[0,1]^n$= environmental state/loss vector at time $t$.\\
$s_{1:t}:=s_1+...+s_t$= state/loss (similar for $\ell_t$ and $r_t$).\\
$s_{1:T}^{min}=\min_i\{s_{1:T}^i\}$= loss of Best Expert in Hindsight (BEH).\\
$s_{<t}:=s_1+...+s_{t-1}$= state/loss summary ($s_{<0}=0$).\\
$M(s):=\arg\min_{d\in\D}\{d\scp s\}$= best decision on $s$.\\
$T\in\SetN_0$= total time=step, $t\in\SetN$= current time=step.\\
$k^i\geq 0$= penalization = complexity of expert $i$.\\
$q\in\SetR^n$= random vector with independent exponentially distributed components.\\
$I_t:=\arg\min_{i\in\E}\{s_{<t}^i+{k^i-q^i\over\eta_t}\}$= randomized prediction of FPL.\\
$\ell_t:=E[M(s_{<t}+{k-q\over\eta_t})\scp s_t]$= expected loss at time $t$ of FPL (=$E[s_t^{I_t}]$ for $\D=\E$).\\
$r_t:=E[M(s_{1:t}+{k-q\over\eta_t})\scp s_t]$= expected loss at time $t$ of IFPL. \\
$u_t:=M(s_{<t}+{k-q\over\eta_t})\scp s_t$= actual loss at time $t$ of FPL (=$s_t^{I_t}$ for $\D=\E$).\\

%%%%%%%%%%%%%%%%%%%%%%%%%%%%%%%%%%%%%%%%%%%%%%%%%%%%%%%%%%%%%%%
\section{IFPL bounded by Best Expert in Hindsight}\label{secExpMax}\label{secIFPL}
%%%%%%%%%%%%%%%%%%%%%%%%%%%%%%%%%%%%%%%%%%%%%%%%%%%%%%%%%%%%%%%

In this section we provide tools for comparing the loss of IFPL
to the loss of the best expert in hindsight. The first result
bounds the expected error induced by the exponentially distributed
perturbation.

\begin{lemma}[Maximum of Shifted Exponential Distributions]\label{lemExpMax}
Let $q^1,...,q^n$ be (not necessarily independent) exponentially
distributed random variables, i.e.\ $P[q^i]=\e^{-q^i}$ for
$q^i\geq 0$ and $1\leq i\leq n\leq\infty$, and $k^i\in\SetR$ be
real numbers with $u:=\sum_{i=1}^n\e^{-k^i}$. Then
\bqan
  P[\max_i\{q^i-k^i\}\geq a]
  &=& 1-\prod_{i=1}^n \max\{0,1\!-\!\e^{-a-k^i}\}
  \qmbox{if} q^1,...,q^n \;\mbox{are independent,}
\\
  P[\max_i\{q^i-k^i\}\geq a]
  &\leq& \min\{1,u\,\e^{-a}\},
\\
  E[\max_i\{q^i-k^i\}] &\leq& 1+\ln u.
\eqan
\end{lemma}

\paradot{Proof} Using
\beqn
  P[q^i<a] = \max\{0,1\!-\!\e^{-a}\}\geq 1-\e^{-a}
  \qmbox{and}
  P[q^i\geq a] = \min\{1,\e^{-a}\}\leq \e^{-a},
\eeqn
valid for any $a\in\SetR$, the exact expression for $P[\max]$ in
Lemma~\ref{lemExpMax} follows from
\beqn
  P[\max_i\{q^i-k^i\}<a]
  = P[q^i-k^i<a\;\forall i]
  = \prod_{i=1}^n P[q^i<a+k^i]
  = \prod_{i=1}^n \max\{0,\e^{-a-k^i}\}
\eeqn
where the second equality follows from the independence of the
$q^i$. The bound on $P[\max]$ for any $a\in\SetR$ (including negative $a$) follows
from
\beqn
  P[\max_i\{q^i-k^i\}\geq a]
  = P[\exists i:q^i-k^i\geq a]
  \leq \sum_{i=1}^n P[q^i-k^i\geq a]
  \leq \sum_{i=1}^n \e^{-a-k^i} = u\!\cdot\!\e^{-a}
\eeqn
where the first inequality is the union bound.
Using $E[z]\leq E[\max\{0,z\}]=\int_0^\infty P[\max\{0,z\}\geq
y]dy = \int_0^\infty P[z\geq y]dy$ (valid for any real-valued
random variable $z$) for $z=\max_i\{q^i-k^i\}-\ln u$, this implies
\beqn
  E[\max_i\{q^i-k^i\}-\ln u]
  \leq \int_0^\infty P\big[\max_i\{q^i-k^i\}\geq y+\ln u \big]dy
  \leq \int_0^\infty \e^{-y} dy\ = \ 1,
\eeqn
which proves the bound on $E[\max]$.
\qed

If $n$ is finite, a lower bound $E[\max_i q^i]\geq 0.57721+\ln n$
can be derived, showing that the upper bound on $E[\max]$ is quite
tight (at least) for $k^i=0$ $\forall i$.
The following bound generalizes \citep[Lem.3]{Kalai:03} to
arbitrary weights, establishing a relation between IFPL and
the best expert in hindsight.

\begin{theorem}[IFPL bounded by BEH]\label{thIFPL}
Let $\D\subseteq\SetR^n$, $s_t\in\SetR^n$ for $1\leq t\leq T$
(both $\D$ and $s$ may even have negative components, but we assume that all
required extrema are attained), and $q,k\in\SetR^n$. If
$\eta_t>0$ is decreasing in $t$, then the loss of the infeasible FPL
knowing $s_t$ at time $t$ in advance (l.h.s.) can be bounded in
terms of the best predictor in hindsight (first term on r.h.s.) plus
additive corrections:
\beqn
  \sum_{t=1}^T M(s_{1:t}+{k\!-\!q\over\eta_t})\scp s_t
  \leq \min_{d\in\D}\{d\scp(s_{1:T}+{k\over\eta_T})\}
     + {1\over\eta_T}\max_{d\in\D}\{d\scp(q-k)\}
     - {1\over\eta_T} M(s_{1:T}+{k\over\eta_T})\scp q.
\eeqn
\end{theorem}

Note that if $\D=\E$ (or $\D=\Delta$) and $s_t\geq 0$, then
all extrema in the theorem are attained almost surely. The
same holds for all subsequent extrema in the proof and
throughout the paper.

\paradot{Proof} For notational convenience, let $\eta_0=\infty$ and
$\tilde s\leqt=s\leqt+\frac{k-q}{\eta_t}$. Consider the losses
$\tilde s_t=s_t+(k-q)\big(\frac{1}{\eta_t}-\frac{1}{\eta_{t-1}}\big)$
for the moment. We first show by induction on $T$ that the infeasible
predictor $M(\tilde s_{1:t})$ has zero regret for any loss $\tilde
s$, i.e.\
\beq\label{eqnoregret}
  \sum_{t=1}^T M(\tilde s_{1:t})\scp \tilde s_t \leq M(\tilde s_{1:T})\scp \tilde s_{1:T}.
\eeq
For $T=1$ this is obvious. For the induction step from $T-1$ to $T$
we need to show
\beq\label{eq:noregret1}
  M(\tilde s_{1:T})\scp \tilde s_T \leq M(\tilde s_{1:T})\scp
  \tilde s_{1:T} - M(\tilde s_{<T})\scp \tilde s_{<T}.
\eeq
This follows from $\tilde s_{1:T}=\tilde s_{<T}+\tilde s_T$ and
$M(\tilde s_{1:T})\scp \tilde s_{<T} \geq M(\tilde s_{<T})\scp
\tilde s_{<T}$ by minimality of $M$.
Rearranging terms in (\ref{eqnoregret}), we obtain
\beq\label{eqifpl2}
  \sum_{t=1}^T M(\tilde s_{1:t})\scp s_t
  \ \leq\
  M(\tilde s_{1:T})\scp \tilde s_{1:T}- \sum_{t=1}^T M(\tilde s_{1:t})\scp
  (k-q)\Big(\frac{1}{\eta_t}-\frac{1}{\eta_{t-1}}\Big)
\eeq
Moreover, by minimality of $M$,
\bqa
\label{eqifpl4}
M(\tilde s_{1:T})\scp \tilde s_{1:T} & \leq &
M\Big(s_{1:T}+\frac{k}{\eta_T}\Big)\scp
\Big(s_{1:T}+\frac{k-q}{\eta_T}\Big)\\
\nonumber
& = & \min_{d\in\D}\left\{d\scp(s_{1:T}+{k\over\eta_T})\right\}-
M\Big(s_{1:T}+\frac{k}{\eta_T}\Big)\scp
\frac{q}{\eta_T}
\eqa
holds. Using ${1\over\eta_t}-{1\over\eta_{t-1}}\geq 0$ and again
minimality of $M$, we have
\bqa\label{eqifpl3}
\sum_{t=1}^T
({1\over\eta_t}-{1\over\eta_{t-1}})M(\tilde s_{1:t})\scp(q-k) &
\leq & \sum_{t=1}^T
({1\over\eta_t}-{1\over\eta_{t-1}})M(k-q)\scp(q-k)\\
\nonumber
&  = & {1\over\eta_T}M(k-q)\scp(q-k)
= {1\over\eta_T}\max_{d\in\D}\{d\scp(q-k)\}
\eqa
Inserting (\ref{eqifpl4}) and (\ref{eqifpl3}) back into (\ref{eqifpl2})
we obtain the assertion.
\qed

Assuming $q$ random with $E[q^i]=1$ and taking the expectation in
Theorem~\ref{thIFPL}, the last term reduces to
$-{1\over\eta_T}\sum_{i=1}^n M(s_{1:T}+{k\over\eta_T})^i$.
If $\D\geq 0$, the term is negative and may be dropped. In case of
$\D=\E$ or $\Delta$, the last term is identical to
$-{1\over\eta_T}$ (since $\sum_i d^i=1$) and keeping it improves
the bound.
Furthermore, we need to evaluate the expectation of the second to
last term in Theorem~\ref{thIFPL}, namely
$E[\max_{d\in\D}\{d\scp(q-k)\}]$. For $\D=\E$ and $q$ being
exponentially distributed, using Lemma~\ref{lemExpMax}, the
expectation is bounded by $1+\ln u$. We hence get the following
bound:

\begin{corollary}[IFPL bounded by BEH]\label{corIFPL}
For $\D=\E$ and $\sum_i \e^{-k^i}\leq 1$ and
$P[q^i]=\e^{-q^i}$ for $q\geq 0$ and decreasing $\eta_t>0$, the
expected loss of the infeasible FPL exceeds the loss of expert $i$
by at most $k^i/\eta_T$:
\beqn
  r_{1:T} \;\leq\; s_{1:T}^i + {1\over\eta_T}k^i  \quad\forall i.
\eeqn
\end{corollary}

Theorem~\ref{thIFPL} can be generalized to expert
dependent factorizable $\eta_t\leadsto \eta_t^i=\eta_t\cdot\eta^i$
by scaling $k^i\leadsto k^i/\eta^i$ and $q^i\leadsto q^i/\eta^i$.
Using $E[\max_i\{{q^i-k^i\over\eta^i}\}]\leq
E[\max_i\{q^i-k^i\}]/\min_i\{\eta^i\}$, Corollary~\ref{corIFPL},
generalizes to
\beqn
    E[\sum_{t=1}^T M(s_{1:t}+{k-q\over\eta_t^i})\scp s_t]
    \;\leq\; s_{1:T}^i + {1\over\eta_T^i}k^i + {1\over\eta_T^{min}}
    \quad\forall i,
\eeqn
where $\eta_T^{min}:=\min_i\{\eta_T^i\}$.
For example, for $\eta_t^i=\sqrt{k^i/t}$
we get the desired bound $s_{1:T}^i+\sqrt{T\cdot(k^i+4)}$.
Unfortunately we were not able to generalize Theorem~\ref{thFIFPL}
to expert-dependent $\eta$, necessary for the final bound on FPL.
In Section~\ref{secHierarchy} we solve this problem by a hierarchy
of experts.

%%%%%%%%%%%%%%%%%%%%%%%%%%%%%%%%%%%%%%%%%%%%%%%%%%%%%%%%%%%%%%%
\section{Feasible FPL bounded by Infeasible FPL}\label{secFFPL}
%%%%%%%%%%%%%%%%%%%%%%%%%%%%%%%%%%%%%%%%%%%%%%%%%%%%%%%%%%%%%%%

This section establishes the relation between the FPL and IFPL
losses. Recall that $\ell_t=E\big[M(s_{<t}+{k-q\over\eta_t})\scp
s_t\big]$ is the expected loss of FPL at time $t$ and
$r_t=E\big[M(s_{1:t}+{k-q\over\eta_t})\scp s_t\big]$ is the
expected loss of IFPL at time $t$.

\begin{theorem}[FPL bounded by IFPL]\label{thFIFPL}
For $\D=\E$ and $0\leq s_t^i\leq 1$ $\forall i$ and arbitrary
$s_{<t}$ and $P[q]=\e^{-\sum_i q^i}$ for $q\geq 0$, the expected
loss of the feasible FPL is at most a factor $\e^{\eta_t}>1$
larger than for the infeasible FPL:
\beqn
  \ell_t\leq \e^{\eta_t}r_t, \qmbox{which implies}
  \ell_{1:T}-r_{1:T}\leq \sum_{t=1}^T\eta_t \ell_t.
\eeqn
Furthermore, if $\eta_t\leq 1$, then also $\ell_t\leq
(1+\eta_t+\eta_t^2)r_t\leq (1+2\eta_t)r_t$.
\end{theorem}

\paradot{Proof}
Let $s=s_{<t}+\sooe k$ be the past cumulative penalized state
vector, $q$ be a vector of independent exponential distributions,
i.e.\ $P[q^i]=\e^{-q^i}$, and $\eta=\eta_t$.
Then
\beqn
  {P[q^j\geq \eta(s^j-m+1)]\over P[q^j\geq\eta(s^j-m)]}
  = \left\{%
\begin{array}{ccc}
  \e^{-\eta}        & \mbox{if} & s^j\geq m \\
  \e^{-\eta(s^j-m+1)} & \mbox{if} & m-1\leq s^j\leq m \\
  1                  & \mbox{if} & s^j\leq m-1 \\
\end{array}%
\right\} \geq \e^{-\eta}
\eeqn
We now define the random variables $I:=\arg\min_i\{s^i-\sooe q^i\}$ and
$J:=\arg\min_i\{s^i+s_t^i-\sooe q^i\}$, where $0\leq s_t^i\leq 1$
$\forall i$. Furthermore, for fixed vector $x\in\SetR^n$ and fixed
$j$ we define $m:=\min_{i\neq j}\{s^i-\sooe x^i\}\leq \min_{i\neq
j}\{s^i+s_t^i-\sooe x^i\}=:m'$.
With this notation and using the independence of $q^j$ from $q^i$
for all $i\neq j$, we get
\beqn
  P[I=j|q^i=x^i\,\forall i\neq j]
  \;=\; P[s^j-\sooe q^j\leq m|q^i=x^i\,\forall i\neq j]
  \;=\; P[q^j\geq\eta(s^j-m)]
\eeqn
\beqn
  \;\leq\; \e^\eta P[q^j\geq\eta(s^j-m+1)]
  \;\leq\; \e^\eta P[q^j\geq\eta(s^j+s_t^j-m')]
\eeqn
\beqn
  \;=\; \e^\eta P[s^j+s_t^j-\sooe q^j\leq m'|q^i=x^i\,\forall i\neq j]
  \;=\; \e^\eta P[J=j|q^i=x^i\,\forall i\neq j]
\eeqn
Since this bound holds under any condition $x$, it also holds
unconditionally, i.e.\ $P[I=j]\leq \e^\eta P[J=j]$. For
$\D=\E$ we have $s_t^I=M(s_{<t}+{k-q\over\eta})\scp s_t$ and
$s_t^J=M(s_{1:t}+{k-q\over\eta})\scp s_t$, which implies
\beqn
  \ell_t
  \;=\;E[s_t^I]
  \;=\; \sum_{j=1}^n s_t^j\!\cdot\!P[I=j]
  \;\leq\; \e^\eta \sum_{j=1}^n s_t^j\!\cdot\!P[J=j]
  \;=\; \e^\eta E[s_t^J]
  \;=\; \e^\eta r_t.
\eeqn
Finally, $\ell_t-r_t\leq\eta_t\ell_t$ follows from $r_t\geq
\e^{-\eta_t}\ell_t\geq (1-\eta_t)\ell_t$, and $\ell_t\leq
\e^{\eta_t}r_t\leq (1+\eta_t+\eta_t^2)r_t\leq (1+2\eta_t)r_t$ for
$\eta_t\leq 1$ is elementary.
\qed

\paradot{Remark}
As done by \citet{Kalai:03}, one can prove a similar statement
for general decision space $\D$ as long as
$\sum_i|s_t^i|\leq A$ is guaranteed for some $A>0$: In this
case, we have $\ell_t\leq \e^{\eta_t A}r_t$. If $n$ is finite,
then the bound holds for $A=n$. For $n=\infty$, the assertion
holds under the somewhat unnatural assumption that $\S$ is
$l^1$-bounded.

%%%%%%%%%%%%%%%%%%%%%%%%%%%%%%%%%%%%%%%%%%%%%%%%%%%%%%%%%%%%%%%
\section{\boldmath Combination of Bounds and Choices for $\eta_t$}\label{secBounds}
%%%%%%%%%%%%%%%%%%%%%%%%%%%%%%%%%%%%%%%%%%%%%%%%%%%%%%%%%%%%%%%

Throughout this section, we assume
\beq
\label{eq:Assumptions}
  \D=\E,\quad s_t\in[0,1]^n\ \forall t,\quad
  P[q]=\e^{-\sum_i q^i} \;\mbox{for}\; q\geq 0,\ \qmbox{and}
  \sum_i \e^{-k^i}\leq 1.
\eeq
We distinguish \emph{static} and \emph{dynamic} bounds. Static
bounds refer to a constant $\eta_t\equiv\eta$. Since this value
has to be chosen in advance, a static choice of $\eta_t$ requires
certain prior information and therefore is not practical in many
cases. However, the static bounds are very easy to derive, and
they provide a good means to compare different PEA algorithms. If
on the other hand the algorithm shall be applied without
appropriate prior knowledge, a dynamic choice of $\eta_t$ depending
only on $t$ and/or past observations, is necessary.

\begin{theorem}[FPL bound for static $\eta_t=\eta\propto 1/\sqrt{L}$]\label{thFPLStatic}
Assume (\ref{eq:Assumptions}) holds, then the expected loss
$\ell_t$ of feasible FPL, which employs the prediction of the
expert $i$ minimizing $s_{<t}^i+{k^i-q^i\over\eta_t}$, is bounded
by the loss of the best expert in hindsight in the following way:
\bqan
  i) & & \nq
  \mbox{For}\quad \eta_t=\eta=1/\sqrt{L}
  \qmbox{with} L\geq\ell_{1:T}
  \qmbox{we have}
\\
     & & \nq
  \ell_{1:T}
  \;\leq\; s_{1:T}^i + \sqrt{L}(k^i+1) \quad\forall i
\\
  ii) & & \nq
  \mbox{For}\quad \eta_t=\sqrt{K/L}
  \qmbox{with} L\geq\ell_{1:T}
  \qmbox{and} k^i\leq K \;\forall i
  \qmbox{we have}
\\
    & & \nq
  \ell_{1:T}
  \;\leq\; s_{1:T}^i + 2\sqrt{LK} \quad\forall i
\\
  iii) & & \nq
  \mbox{For}\quad \eta_t=\sqrt{k^i/L}
  \qmbox{with} L\geq \max\{s_{1:T}^i,k^i\}
  \qmbox{we have}
\\
    & & \nq
  \ell_{1:T}
  \;\leq\; s_{1:T}^i + 2\sqrt{Lk^i}+3k^i
\eqan
\end{theorem}

Note that according to assertion $(iii)$, knowledge of only the
\emph{ratio} of the complexity and the loss of the best
expert is sufficient in order to obtain good static bounds, even
for non-uniform complexities.

\paradot{Proof} $(i,ii)$ For $\eta_t=\sqrt{K/L}$ and $L\geq\ell_{1:T}$,
from Theorem~\ref{thFIFPL} and Corollary
\ref{corIFPL}, we get
\beqn
  \ell_{1:T}-r_{1:T}
  \leq \sum_{t=1}^T\eta_t\ell_t
  = \ell_{1:T}\sqrt{K/L}\leq\sqrt{LK}
  \qmbox{and}
  r_{1:T}-s_{1:T}^i
  \leq k^i/\eta_T=k^i\sqrt{L/K}
\eeqn
Combining both, we get
$\ell_{1:T}-s_{1:T}^i\leq\sqrt{L}(\sqrt{K}+k^i/\sqrt{K})$.
$(i)$ follows from $K=1$ and $(ii)$ from $k^i\leq K$.

\noindent
$(iii)$ For $\eta=\sqrt{k^i/L}\leq 1$ we get
\bqan
  \ell_{1:T}
  & \leq & \e^\eta r_{1:T}
  \leq (1+\eta+\eta^2)r_{1:T}
  \leq (1+\sqrt{k^i\over L}+{k^i\over L})(s_{1:T}^i+\sqrt{L\over
  k^i}k^i)\\
  & \leq & s_{1:T}^i+\sqrt{Lk^i} +(\sqrt{k^i\over L}+{k^i\over L})(L+\sqrt{Lk^i})
  = s_{1:T}^i + 2\sqrt{Lk^i} +(2+\sqrt{k^i\over L})k^i
\eqan
\qed

The static bounds require knowledge of an upper bound $L$ on the
loss (or the ratio of the complexity of the best expert and its
loss). Since the instantaneous loss is bounded by $1$, one may set
$L=T$ if $T$ is known in advance. For finite $n$ and $k^i=K=\ln
n$, bound $(ii)$ gives the classic regret $\propto\sqrt{T\ln
n}$. If neither $T$ nor $L$ is known, a dynamic choice of $\eta_t$
is necessary. We first present bounds with regret $\propto\sqrt{T}$,
thereafter with regret $\propto\sqrt{s_{1:T}^i}$.

\begin{theorem}[FPL bound for dynamic $\eta_t\propto 1/\sqrt{t}$]\label{thFPLTDynamic}
Assume (\ref{eq:Assumptions}) holds.
\bqan
  i) & & \nq
  \mbox{For}\quad \eta_t=1/\sqrt{t}
  \qmbox{we have}
  \ell_{1:T} \;\leq\; s_{1:T}^i + \sqrt{T}(k^i+2) \quad\forall i
\\
  ii) & & \nq
  \mbox{For}\quad \eta_t=\sqrt{K/2t}
  \;\;\mbox{and}\;\; k^i\leq K \;\forall i
  \;\;\mbox{we have}\;\;
  \ell_{1:T} \;\leq\; s_{1:T}^i + 2\sqrt{2TK}
  \quad\forall i
\eqan
\end{theorem}

\paradot{Proof} For $\eta_t=\sqrt{K/2t}$, using
$\sum_{t=1}^T{1\over\sqrt{t}}\leq\int_0^T{dt\over\sqrt{t}}=
2\sqrt{T}$ and $\ell_t\leq 1$ we get
\beqn
  \ell_{1:T}-r_{1:T}
  \leq \sum_{t=1}^T \eta_t
  \leq \sqrt{2TK}
  \qmbox{and}
  r_{1:T}-s_{1:T}^i
  \leq {k^i/\eta_T}=k^i\sqrt{2T\over K}
\eeqn
Combining both, we get
$\ell_{1:T}-s_{1:T}^i \leq \sqrt{2T}(\sqrt{K}+k^i/\sqrt{K})$.
$(i)$ follows from $K=2$ and $(ii)$ from $k^i\leq K$.
\qed

In Theorem~\ref{thFPLStatic} we assumed knowledge of an
upper bound $L$ on $\ell_{1:T}$. In an adaptive form,
$L_t:=\ell_{<t}+1$, known at the beginning of time $t$, could be used
as an upper bound on $\ell_{1:t}$ with corresponding adaptive
$\eta_t\propto 1/\sqrt{L_t}$. Such choice of $\eta_t$ is also
called \emph{self-confident} \citep{Auer:02pea}.

\begin{theorem}[FPL bound for self-confident $\eta_t\propto 1/\sqrt{\ell_{<t}}$]\label{thFPLLDynamic}
Assume (\ref{eq:Assumptions}) holds.
\bqan
  i) & & \nq
  \mbox{For}\quad \eta_t=1/\sqrt{2(\ell_{<t}+1)}
  \qmbox{we have}
\\
   & & \nq
  \ell_{1:T}
  \;\leq\; s_{1:T}^i + (k^i\!+\!1)\sqrt{2(s_{1:T}^i\!+\!1)} + 2(k^i\!+\!1)^2
  \quad\forall i
\\
  ii) & & \nq
  \mbox{For}\quad \eta_t=\sqrt{K/2(\ell_{<t}+1)}
  \qmbox{and} k^i\leq K \;\forall i
  \qmbox{we have}
\\
    & & \nq
  \ell_{1:T}
  \;\leq\; s_{1:T}^i + 2\sqrt{2(s_{1:T}^i\!+\!1)K} + 8K
  \quad\forall i
\eqan
\end{theorem}

\paradot{Proof} Using
$\eta_t=\sqrt{K/2(\ell_{<t}+1)}\leq\sqrt{K/2\ell_{1:t}}$ and
${b-a\over\sqrt
b}=(\sqrt{b}-\sqrt{a})(\sqrt{b}+\sqrt{a}){1\over\sqrt{b}}\leq
2(\sqrt{b}-\sqrt{a})$ for $a\leq b$ and $t_0:=\min\{t:\ell_{1:t}>0\}$ we get
\beqn\label{eqLD}
  \ell_{1:T}\!-\!r_{1:T}
  \leq \sum_{t=t_0}^T \eta_t\ell_t
  \leq \sqrt{K\over 2}\sum_{t=t_0}^T {\ell_{1:t}\!-\!\ell_{<t}\over\sqrt{\ell_{1:t}}}
  \leq \sqrt{2K}\sum_{t=t_0}^T [\sqrt{\ell_{1:t\!\!}}\,-\!\sqrt{\ell_{<t\!\!}}\;]
  = \sqrt{2K}\sqrt{\ell_{1:T}}
\eeqn
Adding
$r_{1:T}-s_{1:T}^i \leq {k^i\over\eta_T} \leq
k^i\sqrt{2(\ell_{1:T}+1)/K}$ we get
\beqn
  \ell_{1:T}-s_{1:T}^i
  \leq \sqrt{2\bar\kappa^i(\ell_{1:T}\!+\!1)},
  \qmbox{where}
  \sqrt{\bar\kappa^i}:=\sqrt{K}+k^i/\sqrt{K}.
\eeqn
Taking the square and solving the resulting quadratic inequality
w.r.t.\ $\ell_{1:T}$ we get
\beqn
  \ell_{1:T}
  \leq s_{1:T}^i + \bar\kappa^i + \sqrt{2(s_{1:T}^i\!+\!1)\bar\kappa^i+(\bar\kappa^i)^2}
  \leq s_{1:T}^i + \sqrt{2(s_{1:T}^i\!+\!1)\bar\kappa^i} + 2\bar\kappa^i
\eeqn
For $K=1$ we get $\sqrt{\bar\kappa^i}=k^i+1$ which yields $(i)$.
For $k^i\leq K$ we get $\bar\kappa^i\leq 4K$ which yields $(ii)$.
\qed

The proofs of results similar to $(ii)$ for WM for 0/1 loss
all fill several pages \citep{Auer:02pea,Yaroshinsky:04}. The
next result establishes a similar bound, but instead of using
the \emph{expected} value $\ell\ltt$, the \emph{best loss so
far}
$s\ltt\smin$ is used. This may have computational advantages,
since $s\ltt\smin$ is immediately available, while $\ell\ltt$
needs to be evaluated (see discussion in Section~\ref{secMisc}).

\begin{theorem}[FPL bound for adaptive $\eta_t\propto 1/\sqrt{s\ltt\smin}$]\label{thFPL2}
Assume (\ref{eq:Assumptions}) holds.
\bqan
  i) & & \nq
  \mbox{For}\quad \eta_t = 1/\min_i\{k^i+\sqrt{(k^i)^2+2s^i\ltt+2}\}
  \qmbox{we have}
\\
   & & \nq
  \ell\leqT \;\leq\; s\leqT^i+(k^i\!+2)\sqrt{2s\leqT^i}+2(k^i\!+2)^2
  \quad \forall i
\\
  ii) & & \nq
  \mbox{For}\quad \eta_t =
  \sqrt{\odt}\!\cdot\!\min\{1,\sqrt{K/s\ltt\smin}\}
  \qmbox{and} k^i\leq K \;\forall i
  \qmbox{we have}
\\
    & & \nq
  \ell\leqT \;\leq\;
  s\leqT^i+2\sqrt{2K s\leqT^i}+5K\ln(s\leqT^i)+3K+6
  \quad \forall i
\eqan
\end{theorem}
We briefly motivate the strange looking choice for $\eta_t$ in
$(i)$. The first naive candidate, $\eta_t\propto 1/\sqrt{s\ltt^{min}}$,
turns out too large. The next natural trial is requesting
$\eta_t=1/\sqrt{2\min\{s\ltt^i+\frac{k^i}{\eta_t}\}}$. Solving
this equation results in $\eta_t=1/(k^i+\sqrt{(k^i)^2+2s\ltt^i})$,
where $i$ be the index for which $s\ltt^i+\frac{k^i}{\eta_t}$ is
minimal.

\paradot{Proof}
Define the minimum of a vector as its minimum component, e.g.\
$\min(k)=k\smin$.
For notational convenience, let
$\eta_0=\infty$ and $\tilde s\leqt=s\leqt+\frac{k-q}{\eta_t}$.
Like in the proof of Theorem~\ref{thIFPL}, we consider one
exponentially distributed perturbation $q$. Since $M(\tilde
s\leqt)\scp \tilde s_t \leq M(\tilde s\leqt)\scp \tilde s\leqt-
M(\tilde s\ltt)\scp \tilde s\ltt$ by (\ref{eq:noregret1}), we have
\beqn
M(\tilde s\leqt)\scp s_t\leq M(\tilde s\leqt)\scp
\tilde s\leqt- M(\tilde s\ltt)\scp \tilde s\ltt -
M(\tilde s\leqt)\scp
\left(\frac{k-q}{\eta_{t}}-\frac{k-q}{\eta_{t-1}}\right)
\eeqn
Since $\eta_t\leq\sqrt{^1\!/_2}$, Theorem~\ref{thFIFPL} asserts
$\ell_t\leq E[(1+\eta_t+\eta_t^2)M(\tilde s\leqt)\scp s_t]$, thus
$\ell\leqT\leq A+B$, where
\bqan
A & = & \sum_{t=1}^T E\left[(1+\eta_t+\eta_t^2)(M(\tilde
s\leqt)\scp
\tilde s\leqt- M(\tilde s\ltt)\scp \tilde s\ltt)\right]\\
& = &
E[(1+\eta_T+\eta_T^2)M(\tilde s\leqT) \scp\tilde s\leqT]
- E[(1+\eta_1+\eta_1^2)\min(\frac{k-q}{\eta_1})]\\
&& + \sum_{t=1}^{T-1}E\left[
(\eta_t-\eta_{t+1}+\eta_t^2-\eta_{t+1}^2)M(\tilde
s\leqt)\scp\tilde s\leqt\right]
\mbox{\quad and}\\
B & = & \sum_{t=1}^T E\left[(1+\eta_t+\eta_t^2) M(\tilde
s\leqt)\scp
\left(\frac{q-k}{\eta_{t}}-\frac{q-k}{\eta_{t-1}}\right)\right]\\
& \leq & \sum_{t=1}^T (1+\eta_t+\eta_t^2)
\left(\frac{1}{\eta_{t}}-\frac{1}{\eta_{t-1}}\right)
=\frac{1+\eta_T+\eta_T^2}{\eta_T}+
\sum_{t=1}^{T-1}\frac{\eta_t-\eta_{t+1}+\eta_t^2-\eta_{t+1}^2}{\eta_t}
\eqan
Here, the estimate for $B$ follows from
$\frac{1}{\eta_t}-\frac{1}{\eta_{t-1}}\geq 0$ and
$E [M(\eta_t s\leqt+k-q)\scp(q-k)]\leq E[\max_i\{q^i-k^i\}]\leq 1$, which
in turn holds by minimality of $M$, $\sum_i \e^{-k^i}\leq 1$ and
Lemma~\ref{lemExpMax}. In order to estimate $A$, we set
$\bar s\leqt=s\leqt+\frac{k}{\eta_t}$ and observe
$M(\tilde s\leqt)\scp\tilde s\leqt\leq
M(\bar s\leqt)\scp(\bar s\leqt-\frac{q}{\eta_t})$ by minimality
of $M$. The expectations of $q$ can then be evaluated to
$E[M(\bar s\leqt)\scp q]=1$, and as before we have $E[-\min(k-q)]\leq 1$.
Hence
\bqa
\nonumber
\ell\leqT & \leq & A+B\ \leq\
(1+\eta_T+\eta_T^2)\left(M(\bar s\leqT)\scp\bar
s\leqT-\frac{1}{\eta_T}\right)
+ \frac{1+\eta_1+\eta_1^2}{\eta_1}\\
\label{eq:basicest}
&& + \sum_{t=1}^{T-1}
(\eta_t-\eta_{t+1}+\eta_t^2-\eta_{t+1}^2)\left(M(\bar
s\leqt)\scp\bar s\leqt-\frac{1}{\eta_t}\right)+B\\
\nonumber
& \leq &
(1+\eta_T+\eta_T^2)\min(\bar s\leqT)+
\sum_{t=1}^{T-1} (\eta_t-\eta_{t+1}+\eta_t^2-\eta_{t+1}^2)
\min(\bar s\leqt)+\frac{1}{\eta_1}+2.
\eqa
We now proceed by considering the two parts of the theorem
separately.

$(i)$
Here,
$\eta_t=1/\min(k+\sqrt{k^2+2s\ltt+2})$. Fix $t\leq T$ and
choose $m$ such that
$k^m+\sqrt{(k^m)^2+2s\ltt^m+2}$ is minimal. Then
\beqn \min(s\leqt+\frac{k}{\eta_t})
\leq s\ltt^m+1+\frac{k^m}{\eta_t}
=
\mbox{$\frac{1}{2}$}\big(k^m+\sqrt{(k^m)^2+2s\ltt^m+2}\big)^2=\frac{1}{2\eta_t^2}
\leq\frac{1}{2\eta_t\eta_{t+1}}.
\eeqn
We may overestimate the quadratic terms $\eta_t^2$ in
(\ref{eq:basicest}) by $\eta_t$ -- the easiest justification
is that we could have started with the cruder estimate
$\ell_t\leq(1+2\eta_t)r_t$ from Theorem~\ref{thFIFPL}. Then
\bqan
\ell\leqT & \leq &
(1+2\eta_T)\min(s\leqT+\frac{k}{\eta_T})+ 2\sum_{t=1}^{T-1}
(\eta_t-\eta_{t+1})\min(s\leqt+\frac{k}{\eta_t})+\frac{1}{\eta_1}+2\\
& \leq &
(1+2\eta_T)\frac{1}{2\eta_T^2}+ 2\sum_{t=1}^{T-1}
(\eta_t-\eta_{t+1})\frac{1}{2\eta_t^2}+\frac{1}{\eta_1}+2\\
& \leq & \frac{1}{2\eta_T^2}+\frac{1}{\eta_T}+
\sum_{t=1}^{T-1}\left(\frac{1}{\eta_{t+1}}-\frac{1}{\eta_t}\right)+\frac{1}{\eta_1}+2\\
& \leq &
\mbox{$\frac{1}{2}$}\min(k+\sqrt{k^2+2s\ltT+2})^2+2\min(k+\sqrt{k^2+2s\ltT+2}) +2\\
& \leq &
s\leqT^i+(k^i+2)\sqrt{2s\leqT^i}+2(k^i)^2+6k^i+6
\quad\mbox{for all}\ i.
\eqan
This proves the first part of the theorem.

$(ii)$ Here we have $K\geq k^i$ for all $i$. Abbreviate
$a_t=\max\{K,s\leqt\smin\}$ for $1\leq t\leq T$, then
$\eta_t=\sqrt{\frac{K}{2a_{t-1}}}$,
$a_t\geq K$, and $a_t-a_{t-1}\leq 1$ for all $t$. Observe
$M(\bar s\leqt)=M(s\leqt)$,
$\eta_t-\eta_{t+1}=\frac{\sqrt K(a_t-a_{t-1})}
{\sqrt 2\sqrt{a_t}\sqrt{a_{t-1}} (\sqrt{a_t}+\sqrt{a_{t-1}})}$,
$\eta_t^2-\eta_{t+1}^2=\frac{K(a_t-a_{t-1})}{2a_ta_{t-1}}$, and
$\frac{a_t-a_{t-1}} {2 a_{t-1}}\leq
\ln(1+\frac{a_t-a_{t-1}}{a_{t-1}})=\ln (a_t)-\ln (a_{t-1})$ which is true for
$\frac{a_t-a_{t-1}}{a_{t-1}}\leq\frac{1}{K}\leq\frac{1}{\ln 2}$. This
implies
\bqan
\frac{(\eta_t-\eta_{t+1})K}{\eta_t} & \leq &
\frac{K(a_t-a_{t-1})}
{2 a_{t-1}}\leq K\ln\left(1+\frac{a_t-a_{t-1}}{a_{t-1}}\right)
= K\big(\ln(a_t)-\ln(a_{t-1})\big), \\
(\eta_t-\eta_{t+1})s\leqt\smin & \leq &
\frac{\sqrt K(a_t-a_{t-1})(\sqrt{a_{t-1}}+
\sqrt{a_t}-\sqrt{a_{t-1}})}
{\sqrt 2\sqrt{a_{t-1}} (\sqrt{a_t}+\sqrt{a_{t-1}})}\\
& = & \sqrt{\frac{K}{2}}(\sqrt{a_t}-\sqrt{a_{t-1}})
+\frac{\sqrt K(a_t-a_{t-1})^2}{\sqrt{2a_{t-1}}(\sqrt{a_t}+\sqrt{a_{t-1}})^2}\\
& \leqs
{\hspace*{-5cm}\rlap{\fbox{$\stackrel{\mbox{\tiny use
}\scriptstyle a_t-a_{t-1}\leq 1} {\mbox{\tiny and }
\scriptstyle a_{t-1}\geq K}$}}}
& \sqrt{\frac{K}{2}}(\sqrt{a_t}-\sqrt{a_{t-1}})+
\frac{1}{2\sqrt 2}\big(\ln(a_t)-\ln(a_{t-1})\big),\\
\frac{(\eta_t^2-\eta_{t+1}^2)K}{\eta_t} & = &
\frac{K\sqrt{K}(a_t-a_{t-1})}{\sqrt{2}a_t\sqrt{a_{t-1}}}
\leqs {\fbox{$\scriptstyle a_{t-1}\geq K$}}
\sqrt{2}K\big(\ln(a_t)-\ln(a_{t-1})\big), \mbox{ and}\\
(\eta_t^2-\eta_{t+1}^2)s\leqt\smin & \leq &
\frac{K(a_t-a_{t-1})}{2a_{t-1}} \leq
K\big(\ln(a_t)-\ln(a_{t-1})\big),
\eqan
The logarithmic estimate in the second and
third bound is unnecessarily rough and for convenience only.
Therefore, the coefficient of the log-term in the final bound of
the theorem can be reduced to
$2K$ without much effort. Plugging the above estimates back into
(\ref{eq:basicest}) yields
\bqan
\ell\leqT & \leq & s\leqT\smin+\sqrt{\frac{K}{2} s\leqT\smin}+\sqrt{2K s\leqT\smin}+3K+2
+\sqrt{\frac{K}{2} s\leqT\smin}+
\big(\mbox{$\frac{7}{2}$}K+\mbox{$\frac{1}{2\sqrt 2}$}\big)\ln(s\leqT\smin)\\
&&+\frac{1}{\eta_1}+2
\leq s\leqT\smin+2\sqrt{2K
s\leqT\smin}+5K\ln(s\leqT\smin)+3K+6.
\eqan
This completes the proof.
\qed

Theorem~\ref{thFPLLDynamic} and Theorem~\ref{thFPL2} $(i)$
immediately imply the following bounds on the
$\sqrt{\mbox{Loss}}$-regrets:
$\sqrt{\ell_{1:T}}\leq\sqrt{s_{1:T}^i+1}+\sqrt{8K}$,
$\sqrt{\ell_{1:T}}\leq\sqrt{s_{1:T}^i+1}+\sqrt{2}(k^i+1)$, and
$\sqrt{\ell_{1:T}}\leq\sqrt{s_{1:T}^i}+\sqrt{2}(k^i+2)$,
respectively.

\paradot{Remark}
The same analysis as for Theorems
[\ref{thFPLStatic}--\ref{thFPL2}]$(ii)$ applies to general $\D$,
using $\ell_t\leq \e^{\eta_t n}r_t$ instead of $\ell_t\leq
\e^{\eta_t}r_t$, and leading to an additional factor $\sqrt{n}$ in
the regret. Compare the remark at the end of Section~\ref{secFFPL}.

%%%%%%%%%%%%%%%%%%%%%%%%%%%%%%%%%%%%%%%%%%%%%%%%%%%%%%%%%%%%%%%
\section{Hierarchy of Experts}\label{secHierarchy}
%%%%%%%%%%%%%%%%%%%%%%%%%%%%%%%%%%%%%%%%%%%%%%%%%%%%%%%%%%%%%%%

We derived bounds which do not need prior knowledge of $L$ with
regret $\propto\sqrt{TK}$ and $\propto\sqrt{s_{1:T}^i K}$ for a
finite number of experts with equal penalty $K=k^i=\ln n$. For
an infinite number of experts, unbounded expert-dependent complexity
penalties $k^i$ are necessary (due to constraint $\sum_i
\e^{-k^i}\leq 1$). Bounds for this case (without prior knowledge of
$T$) with regret $\propto k^i\sqrt{T}$ and $\propto
k^i\sqrt{s_{1:T}^i}$ have been derived. In this case, the
complexity $k^i$ is no longer under the square root. Although
this already implies Hannan consistency, i.e.\ the average per
round regret tends to zero as $t\to\infty$, improved regret
bounds $\propto\sqrt{Tk^i}$ and $\propto\sqrt{s_{1:T}^i k^i}$
are desirable and likely to hold. We were not able to derive
such improved bounds for FPL, but for a (slight) modification.
We consider a two-level hierarchy of experts. First consider
an FPL for the subclass of experts of complexity
$K$, for each $K\in\SetN$. Regard these FPL$^K$ as (meta) experts
and use them to form a (meta) FPL. The class of meta experts now
contains for each complexity only one (meta) expert, which allows
us to derive good bounds. In the following, quantities referring
to complexity class $K$ are superscripted by $K$, and meta
quantities are superscripted by $\;\widetilde{}$ .

Consider the class of experts $\E^K:=\{i:K-1<k^i\leq K\}$ of
complexity $K$, for each $K\in\SetN$. FPL$^K$ makes randomized
prediction
$I_t^K:=\arg\min_{i\in\E^K}\{s_{<t}^i+\smash{k^i-q^i\over\eta_t^K}\}$
with $\eta_t^K:=\sqrt{K/2t}$ and suffers loss $u_t^K:=s_t^{I_t^K}$
at time $t$. Since $k^i\leq K$ $\forall i\in\E^k$ we can apply
Theorem~\ref{thFPLTDynamic}$(ii)$ to FPL$^K$:
\beq\label{eqFH}
  E[u_{1:T}^K] \;=\; \ell_{1:T}^K \;\leq\; s_{1:T}^i+ 2\sqrt{2TK}
  \quad \forall i\in\E^K
  \quad \forall K\in\SetN.
\eeq
We now define a meta state $\tilde s_t^K=u_t^K$ and regard FPL$^K$
for $K\in\SetN$ as meta experts, so meta expert $K$ suffers loss
$\tilde s_t^K$. (Assigning expected loss $\tilde
s_t^K=E[u_t^K]=\ell_t^K$ to FPL$^K$ would also work.) Hence the
setting is again an expert setting and we define the meta
$\widetilde{\mbox{FPL}}$ to predict $\tilde
I_t:=\arg\min_{K\in\SetN}\{\tilde s_{<t}^K+{\tilde k^K-\tilde
q^K\over \tilde\eta_t}\}$ with $\tilde\eta_t=1/\sqrt{t}$ and
$\tilde k^K=\odt+2\ln K$ (implying $\sum_{K=1}^\infty \e^{-\tilde
k^K}\leq 1$). Note that $\tilde s_{1:t}^K=\tilde s_1^K+...+\tilde
s_t^K= s_1^{I_1^K}+...+s_t^{I_t^K}$ sums over the same meta state
components $K$, but over different components ${I_t^K}$ in normal
state representation.

By Theorem~\ref{thFPLTDynamic}$(i)$ the $\tilde q$-expected loss
of $\widetilde{\mbox{FPL}}$ is bounded by $\tilde s_{1:T}^K +
\sqrt{T}(\tilde k^K+2)$. As this bound holds for all $q$ it also holds
in $q$-expectation. So if we define $\tilde\ell_{1:T}$ to be the
$q$ {\em and} $\tilde q$ expected loss of
$\widetilde{\mbox{FPL}}$, and chain this bound with (\ref{eqFH})
for $i\in\E^K$ we get:
\bqan
  \tilde\ell_{1:T}
  &\leq& E[\tilde s_{1:T}^K + \sqrt{T}(\tilde k^K\!+2)]
   \;=\; \ell_{1:T}^K + \sqrt{T}(\tilde k^K\!+2) \\
  &\leq& s_{1:T}^i+ \sqrt{T}[2\sqrt{2(k^i\!+1)}+\odt+2\ln (k^i\!+1)+2],
\eqan
where we have used $K\leq k^i+1$. This bound is valid for all $i$
and has the desired regret $\propto\sqrt{T k^i}$. Similarly we can
derive regret bounds $\propto\sqrt{s_{1:T}^i k^i}$ by exploiting
that the bounds in Theorems~\ref{thFPLLDynamic} and \ref{thFPL2}
are concave in $s_{1:T}^i$ and using Jensen's inequality.

\begin{theorem}[Hierarchical FPL bound for dynamic $\eta_t$]\label{thHFPL}
The hierarchical $\widetilde{\mbox{FPL}}$ employs at time $t$
the prediction of expert $i_t:=I_t^{\tilde I_t}$, where
\vspace{-0.5ex}\beqn
  I_t^K:=\mathop{\arg\min}_{i:\lceil k^i\rceil=K}
    \Big\{s_{<t}^i+{\textstyle{k^i-q^i\over\eta_t^K}}\Big\}
  \qmbox{and}
  \tilde I_t:=\mathop{\arg\min}_{K\in\SetN}
    \Big\{s_1^{I_1^K}+...+s_{t-1}^{I_{t-1}^K}+
    {\textstyle{{1\over 2}+2\ln\!K -\tilde q^K\over \tilde\eta_t}}\Big\}
  \vspace{-1.5ex}
\eeqn
Under assumptions (\ref{eq:Assumptions}) and independent $P[\tilde q^K]=\e^{-\tilde
q^K}$ $\forall K\in\SetN$, the
expected loss $\tilde\ell_{1:T}=E[s_1^{i_1}+...+s_T^{i_T}]$ of
$\widetilde{\mbox{FPL}}$ is bounded as follows:
\bqan
  a) & & \nq
  \mbox{For}\quad \eta_t^K=\sqrt{K/2t}
  \qmbox{and} \tilde\eta_t=1/\sqrt{t}
  \qmbox{we have}
\\
   & & \nq
  \tilde\ell_{1:T}
  \;\leq\; s_{1:T}^i + 2\sqrt{2Tk^i}\!\cdot\!\big(1+O({\textstyle{\ln k^i\over \sqrt{k^i}}})\big)
  \quad\forall i.
\\
  b) & & \nq
  \mbox{For $\tilde\eta_t$ as in $(i)$ and $\eta_t^K$ as in $(ii)$
  of Theorem $\{{\ref{thFPLLDynamic}\atop\ref{thFPL2}}\}$ we have}
\\
    & & \nq
  \tilde\ell_{1:T}
  \;\leq\; s_{1:T}^i + 2\sqrt{2s_{1:T}^i k^i}\!\cdot\!\big(1+O({\textstyle{\ln k^i\over \sqrt{k^i}}})\big)
  + {\textstyle\big\{{O(k^i)\atop O(k^i\ln s_{1:T}^i)}\big\}}
  \quad\forall i.
\eqan
\end{theorem}
The hierarchical $\widetilde{\mbox{FPL}}$ differs from a
direct FPL over all experts $\E$. One potential way to prove a
bound on direct FPL may be to show (if it holds) that FPL
performs better than $\widetilde{\mbox{FPL}}$, i.e.\ $\ell_{1:T}\leq
\tilde\ell_{1:T}$. Another way may be to suitably generalize
Theorem~\ref{thFIFPL} to expert dependent $\eta$.

%%%%%%%%%%%%%%%%%%%%%%%%%%%%%%%%%%%%%%%%%%%%%%%%%%%%%%%%%%%%%%%
\section{Lower Bound on FPL}\label{secLowFPL}
%%%%%%%%%%%%%%%%%%%%%%%%%%%%%%%%%%%%%%%%%%%%%%%%%%%%%%%%%%%%%%%

A lower bound on FPL similar to the upper bound in Theorem
\ref{thIFPL} can also be proven.

\begin{theorem}[FPL lower-bounded by BEH]\label{thLowFPL}
Let $n$ be finite. Assume $\D\subseteq\SetR^n$ and $s_t\in\SetR^n$
are chosen such that the required extrema exist (possibly
negative), $q\in\SetR^n$, and $\eta_t>0$ is a
decreasing sequence. Then the loss of FPL for uniform
complexities (l.h.s.) can be lower-bounded in terms of the best
predictor in hindsight (first term on r.h.s.) plus/minus additive
corrections:
\beqn
  \sum_{t=1}^T M(s_{<t}-{q\over\eta_t})\scp s_t
  \geq \min_{d\in\D}\{d\scp s_{1:T}\}
     - {1\over\eta_T}\max_{d\in\D}\{d\scp q\}
     + \sum_{t=1}^T ({1\over\eta_t}\!-\!{1\over\eta_{t-1}}) M(s_{<t})\scp q
\eeqn
\end{theorem}

\paradot{Proof}
For notational convenience, let $\eta_0=\infty$ and
$\tilde s\leqt=s\leqt-\frac{q}{\eta_t}$. Consider the losses
$\tilde s_t=s_t-q\big(\frac{1}{\eta_t}-\frac{1}{\eta_{t-1}}\big)$
for the moment. We first show by induction on $T$ that the
predictor $M(\tilde s_{<t})$ has nonnegative regret, i.e.\
\beq\label{eqposregret}
  \sum_{t=1}^T M(\tilde s_{<t})\scp\tilde s_t \geq M(\tilde s_{1:T})\scp\tilde s_{1:T}.
\eeq
For $T=1$ this follows immediately from minimality of $M$
($\tilde s_{<1}:=0$). For the induction step from $T-1$ to $T$ we
need to show
\beqn
  M(\tilde s_{<T})\scp \tilde s_T \geq M(\tilde s_{1:T})\scp \tilde s_{1:T} -
  M(\tilde s_{<T})\scp \tilde s_{<T}.
\eeqn
Due to $\tilde s_{1:T}=\tilde s_{<T}+\tilde s_T$, this is
equivalent to $M(\tilde s_{<T})\scp \tilde s_{1:T} \geq M(\tilde
s_{1:T})\scp \tilde s_{1:T}$, which holds by minimality of
$M$. Rearranging terms in (\ref{eqposregret}) we obtain
\beq\label{eqifpl2l}
  \sum_{t=1}^T M(\tilde s_{<t})\scp s_t
  \geq M(\tilde s_{1:T})\scp \tilde s\leqT
   + \sum_{t=1}^T M(\tilde s_{<t})\scp q
   \Big(\frac{1}{\eta_t}-\frac{1}{\eta_{t-1}}\Big), \quad\mbox{with}
\eeq
\beqn
  M(\tilde s_{1:T})\scp \tilde s\leqT=
  M(s_{1:T}-\frac{q}{\eta_T})\scp s_{1:T}
  -M(s_{1:T}-\frac{q}{\eta_T})\scp \frac{q}{\eta_T}
  \geq \min_{d\in\D}\{d\scp s_{1:T}\}
  - {1\over\eta_T}\max_{d\in\D}\{d\scp q\}
\eeqn
\beqn
\mbox{and}\quad \sum_{t=1}^T M(\tilde s_{<t})\scp q
\Big(\frac{1}{\eta_t}-\frac{1}{\eta_{t-1}}\Big)
\;\geq\; \sum_{t=1}^T \Big({1\over\eta_t}-{1\over\eta_{t-1}}\Big)M(s_{<t})\scp q
\eeqn
Again, the last bound follows from the minimality of $M$, which
asserts that $[M(s-q)-M(s)]\scp s\geq 0\geq
[M(s-q)-M(s)]\scp(s-q)$ and thus implies that $M(s-q)\scp q\geq
M(s)\scp q$. So Theorem \ref{thLowFPL} follows from (\ref{eqifpl2l}).
\qed

Assuming $q$ random with $E[q^i]=1$ and taking the expectation in
Theorem~\ref{thLowFPL}, the last term reduces to
$\sum_t({1\over\eta_t}-{1\over\eta_{t-1}})\sum_i
M(s_{<t})^i$.
If $\D\geq 0$, the term is positive and may be dropped. In case of
$\D=\E$ or $\Delta$, the last term is identical to
${1\over\eta_T}$ (since $\sum_i d^i=1$) and keeping it improves
the bound.
Furthermore, we need to evaluate the expectation of the second to
last term in Theorem~\ref{thLowFPL}, namely
$E[\max_{d\in\D}\{d\scp q\}]$. For $\D=\E$ and $q$ being
exponentially distributed, using Lemma~\ref{lemExpMax} with
$k^i=0$ $\forall i$, the expectation is bounded by $1+\ln n$.
We hence get the following lower bound:

\begin{corollary}[FPL lower-bounded by BEH]\label{corLowFPL}
For $\D=\E$ and any $\S$ and all $k^i$ equal and
$P[q^i]=\e^{-q^i}$ for $q\geq 0$ and decreasing $\eta_t>0$, the
expected loss of FPL is at most
$\ln n/\eta_T$ lower than the loss of the best expert in hindsight:
\beqn
  \ell_{1:T} \;\geq\; s_{1:T}^{min} - {\ln n\over\eta_T}
\eeqn
\end{corollary}

The upper and lower bounds on $\ell_{1:T}$
(Theorem~\ref{thFIFPL} and Corollaries~\ref{corIFPL} and
\ref{corLowFPL}) together show that
\beq\label{eqltos}
  {\ell_{1:t}\over s_{1:t}^{min}} \to 1
  \quad\qmbox{if}\quad
  \eta_t\to 0
  \qmbox{and}
  \eta_t\!\cdot\!s_{1:t}^{min} \to \infty
  \qmbox{and}
  k^i=K\;\forall i
\eeq
For instance, $\eta_t=\sqrt{K/2 s_{<t}^{min}}$. For
$\eta_t=\sqrt{K/2(\ell_{<t}+1)}$ we proved the bound in Theorem
\ref{thFPLLDynamic}$(ii)$. Knowing that $\sqrt{K/2(\ell_{<t}+1)}$
converges to $\sqrt{K/2 s_{<t}^{min}}$ due to (\ref{eqltos}), we
can derive a bound similar to Theorem~\ref{thFPLLDynamic}$(ii)$
for $\eta_t=\sqrt{K/2 s_{<t}^{min}}$. This choice for $\eta_t$ has
the advantage that we do not have to compute $\ell_{<t}$ (cf.\
Section~\ref{secComp}), as also achieved by Theorem~\ref{thFPL2}$(ii)$.

We do not know whether Theorem~\ref{thLowFPL} can be
generalized to expert dependent complexities $k^i$.

%%%%%%%%%%%%%%%%%%%%%%%%%%%%%%%%%%%%%%%%%%%%%%%%%%%%%%%%%%%%%%%
\section{Adaptive Adversary}\label{secAdap}
%%%%%%%%%%%%%%%%%%%%%%%%%%%%%%%%%%%%%%%%%%%%%%%%%%%%%%%%%%%%%%%

In this section we show that bounds that hold against an
oblivious adversary automatically also hold against an
adaptive one.

%-------------------------------%
\paradot{Initial versus independent randomization}
%-------------------------------%
So far we assumed that the perturbations $q$ are sampled only once at
time $t=0$. As already indicated, under the expectation this is
equivalent to generating a new perturbation $q_t$ at each time
step $t$, i.e.\ Theorems \ref{thFIFPL}--\ref{thHFPL} remain valid
for this case. While the former choice was favorable for the
analysis, the latter has two advantages.
First, repeated sampling of the perturbations guarantees better
bounds with high probability (see next section).
Second, if the losses are generated by an adaptive adversary (not
to be confused with an adaptive learning rate) which has access to
FPL's past decisions, then he may after some time figure out the
initial random perturbation and use it to force FPL to have a
large loss.
We now show that the bounds for FPL remain valid, even in case of
an adaptive adversary, if independent randomization $q\leadsto
q_t$ is used.

%-------------------------------%
\paradot{Oblivious versus adaptive adversary}
%-------------------------------%
Recall the protocol for FPL: After each expert $i$ made its
prediction $y_t^i$, and FPL combined them to form its own prediction
$y_t^{\FPL}$, we observe $x_t$, and Loss($x_t,y_t^{\cdots}$) is
revealed for FPL's and each expert's prediction. For independent
randomization, we have $y_t^{\FPL}=y_t^{\FPL}(x_{<t},y_{1:t},q_t)$. For an
oblivious (non-adaptive) adversary, $x_t=x_t(x_{<t},y_{<t})$.
Recursively inserting and eliminating the experts
$y_t^i=y_t^i(x_{<t},y_{<t})$ and $y_t^{\FPL}$, we get the dependencies
\beq\label{eqnAdapDep}
  u_t :=\mbox{Loss}(x_t,y_t^{\FPL}) = u_t(x_{1:t},q_t)
  \qmbox{and}
  s_t^i := \mbox{Loss}(x_t,y_t^i) = s_t^i(x_{1:t}),
\eeq
where $x_{1:t}$ is a ``fixed'' sequence.
With this notation, Theorems \ref{thFPLStatic}--\ref{thFPL2} read
$\ell_{1:T}\equiv E[\sum_{t=1}^T u_t(x_{1:t},q_t)]\leq f(x_{1:T})$
for all $x_{1:T}\in\X^T$, where $f(x_{1:T})$ is one of the
r.h.s.\ in Theorems \ref{thFPLStatic}--\ref{thFPL2}. Noting that
$f$ is independent of $q_{1:T}$, we can write this as
\beq\label{eqnAdapBnd}\label{defAt}
  A_1\leq 0, \qmbox{where}
  A_t(x_{<t},q_{<t})
  := \max_{x_{t:T}}E_{q_{t:T}}\Big[\sum_{\tau=1}^T u_\tau(x_{1:\tau},q_\tau)-f(x_{1:T})\Big],
\eeq
where $E_{q_{t:T}}$ is the expectation w.r.t.\ $q_t...q_T$
(keeping $q_{<t}$ fixed).

For an adaptive adversary, $x_t=x_t(x_{<t},y_{<t},y_{<t}^{\FPL})$
can additionally depend on $y_{<t}^{\FPL}$. Eliminating $y_t^i$
and $y_t^{\FPL}$ we get, again, (\ref{eqnAdapDep}), but
$x_t=x_t(x_{<t},q_{<t})$ is no longer fixed, but an (arbitrary)
random function. So we have to replace $x_t$ by
$x_t(x_{<t},q_{<t})$ in (\ref{eqnAdapBnd}) for $t=1..T$. The
maximization is now a functional maximization over all functions
$x_t(\cdot,\cdot)...x_T(\cdot,\cdot)$. Using ``$\max_{x(\cdot)}E_q
[g(x(q),q)]=E_q\max_x[g(x,q)]$,$\!$'' we can write this as
\beq\label{defBt}
  B_1\stackrel?\leq 0, \qmbox{where}
  B_t(x_{<t},q_{<t})
  := \max_{x_t}E_{q_t}...\max_{x_T}E_{q_T}\Big[\sum_{\tau=1}^T u_\tau(x_{1:\tau},q_\tau)-f(x_{1:T})\Big],
\eeq
So, establishing $B_1\leq 0$ would show that all bounds
also hold in the adaptive case.

\begin{lemma}[Adaptive=Oblivious]\label{lemAdap}
Let $q_1...q_T\in\SetR^T$ be independent random variables,
$E_{q_t}$ be the expectation w.r.t.\ $q_t$, $f$ any function of
$x_{1:T}\in\X^T$, and $u_t$ arbitrary functions of $x_{1:t}$ and $q_t$.
Then, $A_t(x_{<t},q_{<t})=B_t(x_{<t},q_{<t})$ for all $1\leq t\leq T$, where
$A_t$ and $B_t$ are defined in (\ref{defAt}) and (\ref{defBt}).
In particular, $A_1\leq 0$ implies $B_1\leq 0$.
\end{lemma}
\paradot{Proof} We prove $B_t=A_t$ by induction on $t$, which
establishes the theorem. $B_T=A_T$ is obvious. Assume $B_t=A_t$.
Then
\bqan
  B_{t-1} &=& \max_{x_{t-1}}E_{q_{t-1}}B_t \;=\; \max_{x_{t-1}}E_{q_{t-1}} A_t
\\
  &=& \max_{x_{t-1}}E_{q_{t-1}}
      \bigg[\max_{x_{t:T}}E_{q_{t:T}}\Big[\sum_{\tau=1}^T u_\tau(x_{1:\tau},q_\tau)-f(x_{1:T})\Big]\bigg]
\\
  &=& \max_{x_{t-1}}E_{q_{t-1}}
      \bigg[\underbrace{\sum_{\tau=1}^{t-1} u_\tau(x_{1:\tau},q_\tau)}_{\hspace*{-3ex}\text{independent } x_{t:T} \text{ and } q_{t:T}} +
            \underbrace{\max_{x_{t:T}}E_{q_{t:T}}\Big[\sum_{\tau=t}^T u_\tau(x_{1:\tau},q_\tau)-f(x_{1:T})\Big]}_{\text{independent $q_{t-1}$, since the $q_t$ are i.d.}} \bigg]
\\
  &=&
  \max_{x_{t-1}} \bigg[\overbrace{E_{q_{t-1}}\Big[\sum_{\tau=1}^{t-1} u_\tau(x_{1:\tau},q_\tau)\Big]} +
  \overbrace{\max_{x_{t:T}}E_{q_{t:T}}\Big[\sum_{\tau=t}^T u_\tau(x_{1:\tau},q_\tau)-f(x_{1:T})\Big]\bigg]}
\\
  &=&
  \max_{x_{t-1}}\max_{x_{t:T}}E_{q_{t:T}} \bigg[E_{q_{t-1}}\sum_{\tau=1}^{t-1} u_\tau(x_{1:\tau},q_\tau) +
  \sum_{\tau=t}^T u_\tau(x_{1:\tau},q_\tau)-f(x_{1:T})\bigg]
  \;\;=\;\; A_{t-1}
\eqan\qed

\begin{corollary}[FPL Bounds for adaptive adversary]\label{corAdap}
Theorems \ref{thFPLStatic}--\ref{thFPL2} also hold for an adaptive
adversary in case of independent randomization $q\leadsto q_t$.
\end{corollary}

Lemma \ref{lemAdap} shows that every bound of the form
$A_1\leq 0$ proven for an oblivious adversary, implies an
analogous bound
$B_1\leq 0$ for an adaptive adversary. Note that this strong
statement holds only for the \emph{full observation game},
i.e.\ if after each time step we learn all losses. In partial
observation games such as the Bandit case \citep{Auer:95}, our
actual action may depend on our past action by means of our
past observation, and the assertion no longer holds. In this
case, FPL with an adaptive adversary can be analyzed as shown
by \citet{McMahan:04,Poland:05actexp}.
Finally, $y_t^{\IFPL}$ can additionally depend on $x_t$, but the
``reduced'' dependencies (\ref{eqnAdapDep}) are the same as for
FPL, hence, IFPL bounds also hold for adaptive adversary.

%%%%%%%%%%%%%%%%%%%%%%%%%%%%%%%%%%%%%%%%%%%%%%%%%%%%%%%%%%%%%%%
\section{Miscellaneous}\label{secMisc}\label{secComp}
%%%%%%%%%%%%%%%%%%%%%%%%%%%%%%%%%%%%%%%%%%%%%%%%%%%%%%%%%%%%%%%

%-------------------------------%
\paradot{Bounds with high probability}
%-------------------------------%
We have derived several bounds for the expected loss $\ell_{1:T}$
of FPL. The {\em actual} loss at time $t$ is
$u_t=M(s_{<t}+{k-q\over\eta_t})\scp s_t$. A simple Markov inequality shows
that the total actual loss $u_{1:T}$ exceeds
the total expected loss $\ell_{1:T}=E[u_{1:T}]$ by a factor of
$c>1$ with probability at most $1/c$:
\beqn
  P[u_{1:T}\geq c\!\cdot\!\ell_{1:T}]
  \;\leq\; {1/c}.
\eeqn
Randomizing independently for each $t$ as described in the
previous Section, the actual loss is
$u_t=M(s_{<t}+{k-q_t\over\eta_t})\scp s_t$ with the same expected loss
$\ell_{1:T}=E[u_{1:T}]$ as before. The advantage of independent
randomization is that we can get a much better
high-probability bound. We can exploit a Chernoff-Hoeffding
bound \citep[Cor.5.2b]{McDiarmid:89}, valid for arbitrary
independent random variables $0\leq u_t\leq 1$ for
$t=1,...,T$:
\beqn
  P\Big[|u_{1:T}-E[u_{1:T}]|\geq\delta E[u_{1:T}]\Big]
  \;\leq\; 2\exp(-{\textstyle{1\over 3}}\delta^2 E[u_{1:T}]), \qquad 0\leq\delta\leq 1.
\eeqn
For $\delta=\sqrt{3c/\ell_{1:T}}$ we get
\beq\label{eqCH}
  P[|u_{1:T}-\ell_{1:T}|\geq\sqrt{3c\ell_{1:T}}]
  \;\leq\; 2\e^{-c}
  \qmbox{as soon as}
  \ell_{1:T}\geq 3c.
\eeq
Using (\ref{eqCH}), the bounds for $\ell_{1:T}$ of Theorems
\ref{thFPLStatic}--\ref{thFPL2} can be rewritten to yield
similar bounds with high probability ($1-2\e^{-c}$) for $u_{1:T}$
with small extra regret $\propto\sqrt{c\cdot L}$ or $\propto\sqrt{c\cdot
s_{1:T}^i}$.
Furthermore, (\ref{eqCH}) shows that with probability 1,
$u_{1:T}/\ell_{1:T}$ converges rapidly to 1 for
$\ell_{1:T}\to\infty$. Hence we may use the easier to compute
$\eta_t=\sqrt{K/2u_{<t}}$ instead of
$\eta_t=\sqrt{K/2(\ell_{<t}+1)}$, likely with similar bounds on the
regret.

%-------------------------------%
\paradot{Computational Aspects}
%-------------------------------%
It is easy to generate the randomized decision of FPL. Indeed,
only a single initial exponentially distributed vector
$q\in\SetR^n$ is needed. Only for self-confident $\eta_t\propto
1/\sqrt{\ell_{<t}}$ (see Theorem~\ref{thFPLLDynamic}) we need to
compute expectations explicitly. Given $\eta_t$, from $t\leadsto
t+1$ we need to compute $\ell_t$ in order to update $\eta_t$. Note
that $\ell_t=w_t\!\scp s_t$, where $w_t^i=P[I_t=i]$ and
$I_t:=\arg\min_{i\in\E}\{s_{<t}^i+{k^i-q^i\over\eta_t}\}$ is the
actual (randomized) prediction of FPL. With $s:=s_{<t}+k/\eta_t$,
$P[I_t=i]$ has the following representation:
\bqan
  P[I_t=i]
  &=& P[s-{q^i\over\eta_t}\leq s-{q^j\over\eta_t} \;\forall j\neq i] \\
  &=& \int P[s-{q^i\over\eta_t}=m \;\wedge\; s-{q^j\over\eta_t}\geq m \;\forall j\neq i]dm \\
  &=& \int P[q^i=\eta_t(s^i-m)]\cdot\prod_{j\neq i}P[q^j\leq \eta_t(s^j-m)]dm \\[-1ex]
  &=& \int_{-\infty}^{s^{min}} \eta_t \e^{-\eta_t(s^i-m)}
      \prod_{j\neq i}(1-\e^{-\eta_t(s^j-m)})dm \\
  &=& \sum_{{\cal M}:\{i\}\subseteq{\cal M}\subseteq{\cal N}}\!\!
  {\textstyle{(-)^{|{\cal M}|-1}\over|{\cal M}|}}\e^{-\eta_t\sum_{j\in\cal M}(s^j-s^{min})}
\eqan
In the last equality we expanded the product and performed the
resulting exponential integrals. For finite $n$, the
second to last one-dimensional integral should be numerically
feasible. Once the product $\prod_{j=1}^n(1-\e^{-\eta_t(s^j-m)})$
has been computed in time $O(n)$, the argument of the integral can
be computed for each
$i$ in time $O(1)$, hence the overall time to compute $\ell_t$ is
$O(c\cdot n)$, where $c$ is the time to numerically compute one
integral. For infinite $n$, the last sum may be approximated
by the dominant contributions. Alternatively, one can modify
the algorithm by considering only a finite pool of experts in
each time step; see next paragraph. The expectation may also
be approximated by (Monte Carlo) sampling $I_t$ several times.

Recall that approximating $\ell\ltt$ can be avoided by
using $s\ltt\smin$ (Theorem~\ref{thFPL2}) or $u\ltt$ (bounds with
high probability) instead.

%-------------------------------%
\paradot{Finitized expert pool}
%-------------------------------%
In the case of an infinite expert class, FPL has to compute a
minimum over an infinite set in each time step, which is not
directly feasible. One possibility to address this is to
choose the experts from a \emph{finite pool} in each time
step. This is the case in the algorithm of \citet{Gentile:03},
and also discussed by \citet{Littlestone:94}. For FPL, we can
obtain this behavior by introducing an \emph{entering time}
$\tau^i\geq 1$ for each expert. Then expert $i$ is not
considered for $i<\tau^i$. In the bounds, this leads to an
additional $\frac{1}{\eta_T}$ in Theorem \ref{thIFPL} and
Corollary \ref{corIFPL} and a further additional $\tau^i$ in
the final bounds (Theorems \ref{thFPLStatic}--\ref{thFPL2}),
since we must add the regret of the best expert in hindsight
which has already entered the game and the best expert in
hindsight at all. Selecting
$\tau^i=k^i$ implies bounds for FPL with entering times similar to
the ones we derived here. The details and proofs for this
construction can be found in \citep{Poland:05actexp}.

%-------------------------------%
\paradot{Deterministic prediction and absolute loss}
%-------------------------------%
Another use of $w_t$ from the second last paragraph is the following: If
the decision space is $\D=\Delta$, then FPL may make a
deterministic decision $d=w_t\in\Delta$ at time $t$ with bounds
now holding for sure, instead of selecting $e_i$ with probability
$w_t^i$. For example for the absolute loss $s_t^i=|x_t-y_t^i|$
with observation $x_t\in[0,1]$ and predictions $y_t^i\in[0,1]$, a
master algorithm predicting deterministically $w_t\!\scp
y_t\in[0,1]$ suffers absolute loss $|x_t-w_t\!\scp y_t|\leq\sum_i
w_t^i|x_t-y_t^i|=\ell_t$, and hence has the same (or better)
performance guarantees as FPL. In general, masters can be chosen
deterministic if prediction space $\Y$ and loss-function Loss$(x,y)$ are
convex.
For $x_t,y_t^i\in\{0,1\}$, the absolute loss $|x_t-p_t|$ of a master
deterministically predicting $p_t\in[0,1]$ actually coincides with
the $p_t$-expected 0/1 loss of a master predicting 1 with
probability $p_t$. Hence a regret bound for the absolute loss also
implies the same regret for the 0/1 loss.

%%%%%%%%%%%%%%%%%%%%%%%%%%%%%%%%%%%%%%%%%%%%%%%%%%%%%%%%%%%%%%%
\section{Discussion and Open Problems}\label{secConc}
%%%%%%%%%%%%%%%%%%%%%%%%%%%%%%%%%%%%%%%%%%%%%%%%%%%%%%%%%%%%%%%

How does FPL compare with other expert advice algorithms? We
briefly discuss four issues, summarized in Table \ref{tabregconst}.

%-------------------------------%
\paradot{Static bounds}
%-------------------------------%
Here the coefficient of the regret term $\sqrt{KL}$, referred to
as the \emph{leading constant} in the sequel, is $2$ for FPL
(Theorem~\ref{thFPLStatic}). It is thus a factor of $\sqrt 2$
worse than the Hedge bound for arbitrary loss by
\citet{Freund:97}, which is sharp in some sense \citep{Vovk:95}.
This is the price one pays for FPL and its easy analysis for
adaptive learning rate. There is evidence that this (worst-case)
difference really exists and is not only a proof artifact.
%-------------------------------%
%\paradot{Special losses}
%-------------------------------%
For special loss functions, the bounds can sometimes be improved,
e.g.\ to a leading constant of 1 in the static (randomized) WM
case with 0/1 loss \citep{Cesa:97}\footnote{While FPL and Hedge and WMR
\citep{Littlestone:94} can sample an expert without knowing its
prediction, \citet{Cesa:97} need to know the experts' predictions.
Note also that for many (smooth) loss-functions like the quadratic
loss, finite regret can be achieved \citep{Vovk:90}.}.
Because of the structure of the FPL algorithm however, it is
questionable if corresponding bounds hold there.

%-------------------------------%
\paradot{Dynamic bounds}
%-------------------------------%
Not knowing the right learning rate in advance usually costs a
factor of $\sqrt 2$. This is true for Hannan's algorithm
\citep{Kalai:03} as well as in all our cases. Also for binary
prediction with uniform complexities and 0/1 loss, this result
has been established recently -- \citet{Yaroshinsky:04} show a
dynamic regret bound with leading constant $\sqrt 2(1+\eps)$.
Remarkably, the best dynamic bound for a WM variant proven by
\citet{Auer:02pea} has a leading constant $2\sqrt 2$, which
matches ours. Considering the difference in the static case,
we therefore conjecture that a bound with leading constant of
$2$ holds for a dynamic Hedge algorithm.

%-------------------------------%
\paradot{General weights}
%-------------------------------%
While there are several dynamic bounds for uniform weights, the
only previous result for non-uniform weights we know of is
\citep[Cor.16]{Gentile:03}, which gives the dynamic bound
$\ell^{\mbox{\scriptsize Gentile}}\leqT\leq s^i\leqT+i+
O\Big[\sqrt{(s^i\leqT+i)\ln(s^i\leqT+i)}\Big]$ for a $p$-norm
algorithm for the absolute loss. This is comparable to our bound
for rapidly decaying weights $w^i=\exp(-i)$, i.e.\ $k^i=i$. Our
hierarchical FPL bound in Theorem \ref{thHFPL} $(b)$ generalizes
this to arbitrary weights and losses and strengthens it, since
both, asymptotic order and leading constant, are smaller.

It seems that the analysis of all experts algorithms, including
Weighted Majority variants and FPL, gets more complicated for
general weights together with adaptive learning rate, because the
choice of the learning rate must account for both the weight of
the best expert (in hindsight) and its loss. Both quantities are
not known in advance, but may have a different impact on the
learning rate: While increasing the current loss estimate always
decreases $\eta_t$, the optimal learning rate for an expert with
higher complexity would be larger. On the other hand, all analyses
known so far require decreasing $\eta_t$. Nevertheless we
conjecture that the bounds $\propto\sqrt{Tk^i}$ and
$\propto\sqrt{s_{1:T}^i k^i}$ also hold without the hierarchy
trick, probably by using expert dependent learning rate
$\eta_t^i$.

\begin{table}[t]\centering\small
\begin{tabular}{|c|c|c|c|c|}
  \hline
  $\eta$ & Loss & conjecture & Low.Bnd. & Upper Bound \\ \hline
  static & 0/1 & 1            & 1?                         & 1 \citep{Cesa:97} \\
  static & any & $\sqrt{2}$ ! & $\sqrt{2}$ \citep{Vovk:95} & $\sqrt{2}$ \cite[Hedge]{Freund:97}, 2 [FPL] \\
 dynamic & 0/1 & $\sqrt{2}$   & 1? \citep{Hutter:03optisp} & $\sqrt{2}$ \cite{Yaroshinsky:04}, $2\sqrt{2}$ \cite[WM-Type?]{Auer:02pea} \\
 dynamic & any & 2            & $\sqrt{2}$ \citep{Vovk:95} & $2\sqrt{2}$ [FPL], 2 \cite[Bayes]{Hutter:03optisp} \\
  \hline
\end{tabular}
  \caption{\label{tabregconst}Comparison of the constants $c$ in regrets
  $c\sqrt{\mbox{Loss}\times\ln n}$ for various settings and algorithms.}
\end{table}

%-------------------------------%
\paradot{Comparison to Bayesian sequence prediction}
%-------------------------------%
We can also compare the \emph{worst-case} bounds for FPL obtained
in this work to similar bounds for \emph{Bayesian sequence
prediction}. Let $\{\nu_i\}$ be a class of probability
distributions over sequences and assume that the true sequence is
sampled from $\mu\in\{\nu_i\}$ with complexity $k^\mu$ ($\sum_i
\e^{-k^{\nu_i}}\leq 1$). Then it is known that the Bayes optimal
predictor based on the $\e^{-k^{\nu_i}}$-weighted mixture of
$\nu_i$'s has an expected total loss of at most
$L^\mu+2\sqrt{L^\mu k^\mu}+2k^\mu$, where $L^\mu$ is the expected
total loss of the Bayes optimal predictor based on $\mu$
\citep[Thm.2]{Hutter:02spupper},
\citep[Thm.3.48]{Hutter:04uaibook}. Using FPL, we obtained
the same bound except for the leading order constant, but for
any sequence independently of the assumption that it is
generated by $\mu$. This is another indication that a PEA
bound with leading constant 2 could hold. See
\citet{Hutter:04bayespea},
\citet[Sec.6.3]{Hutter:03optisp} and
\citet[Sec.3.7.4]{Hutter:04uaibook} for a more detailed
comparison of Bayes bounds with PEA bounds.

%%%%%%%%%%%%%%%%%%%%%%%%%%%%%%%%%%%%%%%%%%%%%%%%%%%%%%%%%%%%%%%
%         Bibliography        %
%%%%%%%%%%%%%%%%%%%%%%%%%%%%%%%%%%%%%%%%%%%%%%%%%%%%%%%%%%%%%%%

\begin{small}

\end{small}

\end{document}